\documentclass{article}

     \usepackage[final]{neurips_2022}

\usepackage[utf8]{inputenc} 
\usepackage[T1]{fontenc}    
\usepackage{hyperref}       
\usepackage{url}            
\usepackage{booktabs}       
\usepackage{amsfonts}       
\usepackage{nicefrac}       
\usepackage{microtype}      
\usepackage{xcolor}         
\usepackage{graphicx}
\usepackage{subfig}
\usepackage{lscape} 
\usepackage{verbatim}
\usepackage{makecell}
\usepackage[paper=portrait,pagesize]{typearea}

\title{Hierarchical Graph Structures\\for Congestion and ETA Prediction\\ \large Traffic4cast 2022 Competition}

\author{%
  Florian Grötschla\\
  ETH Zurich\\
  \texttt{fgroetschla@ethz.ch} \\
    \And
    Joël Mathys\\
  ETH Zurich\\
  \texttt{jmathys@ethz.ch}} 

\begin{document}

\maketitle

\begin{abstract}
  Traffic4cast is an annual competition to predict spatio temporal traffic based on real world data. 
  We propose an approach using Graph Neural Networks that directly works on the road graph topology which was extracted from OpenStreetMap data. Our architecture can incorporate a hierarchical graph representation to improve the information flow between key intersections of the graph and the shortest paths connecting them.
  Furthermore, we investigate how the road graph can be compacted to ease the flow of information and make use of a multi-task approach to predict congestion classes and ETA simultaneously. 
  Our code and models are released on \href{https://github.com/floriangroetschla/NeurIPS2022-traffic4cast}{Github}.
\end{abstract}

\section{Introduction}

The 2022 edition of the Traffic4cast challenge \citep{traffic4castgithub} focuses on spatio-temporal street congestion predictions 15 minutes into the future based on spatially sparse loop counter data of the past hour. 
In previous editions \citep{traffic4cast21, pmlr-v123-kreil20a, pmlr-v133-kopp21a}, future traffic states had to be predicted on a grid representation of the city. However, in the 2022 competition, cities are given as OpenStreetMap graphs \citep{OpenStreetMap}. Moreover, the loop counter data is given as node features, and the final congestion class predictions are for the edges of the road graph.
One key advantage of loop counter data is that they provide an accurate view of the total traffic in their local proximity. In previous editions, the input was constructed from GPS information, which could only give an approximate picture of the underlying traffic volume. 
In the extended challenge, given the same data, the task is to predict the Expected Time of Arrival (ETA) for provided supersegments throughout the cities. These supersegments are defined as the shortest paths between important key intersections of the individual cities. 

Our approach is centered around Graph Neural Networks (GNNs). Using multiple rounds of message passing, we can directly learn on the graph topology and propagate information between neighboring nodes. 
How far information can be propagated in this setting is a major concern.
Therefore, we focus on building an appropriate architecture that allows information to spread through the entire graph structure. We preprocess the graph to remove artifacts of the OpenStreetMap data. Furthermore, we incorporate a hierarchical representation of the city graph given by the supersegments to improve information flow.  

In the following, we describe our approach in more detail. We outline our data preparation, processing different graph structures and input features. Then, we go into the specific model architectures we used to participate in the core and extended competition as well as the training pipeline. The entire source code and trained models can be found online\footnote{\href{https://github.com/floriangroetschla/NeurIPS2022-traffic4cast}{https://github.com/floriangroetschla/NeurIPS2022-traffic4cast}}.

\section{Model Overview}

We use the same model architecture for the core and extended competition, illustrated in Figure~\ref{fig:model Description}. We prepare the node and edge features for both tasks and load the road and supersegment graph, respectively. The model uses two separate encoders to map the node and edge features to an embedding space. Then, we use several layers of message-passing graph convolutions that propagate information through the graph. 
Each graph convolution takes a set of nodes and edges. For each edge, a message is generated by concatenating the adjacent node embeddings and, if available, the edge embedding. Note that the edges of the original graph all have edge features, whereas the edges introduced by supergraphs in Section \ref{sec:supersegment} do not have any edge features. The message is then passed through an MLP (MessageNN). In the following aggregation step, all messages incident to a node are summed up, concatenated with the original node embedding, and passed through another MLP (UpdateNN). Afterward, we add the previous node embedding with the embedding computed after the aggregation.
The GNN executes six rounds of graph convolutions. The first four rounds (GNNConv) work on the topology of the original road graph. Then, we execute a graph convolution on an extended version of the graph that includes supersegment information (GNNPoolConv). We refer to Section~\ref{sec:supersegment} for an in-depth description of the different approaches. Afterward, we perform two additional rounds of graph convolutions (GNNConv) to derive the final node embeddings. The main difference between GNNConv and GNNPoolConv is that GNNConv includes the edge features of the original graph. In contrast, GNNPoolConv is intended to propagate information on supergraphs that do not have edge features. 

To compute the final predictions for the edges, we concatenate the edge embedding and the node embeddings from before and after the GNN convolutions. This is then passed through a predictor network that generates logits for the different congestion classes. Similarly, a separate predictor is trained to get the ETA estimates for the supersegments. The inputs for this predictor are the node embeddings of the endpoints before and after the GNN convolutions and node embeddings for additional supersegment nodes that we get from one of our supersegment graph constructions. Note that the ETA model is trained in a multi-task manner to predict both the congestion classes of edges and the ETA of supersegments simultaneously. 

\section{Graph Compaction}
\begin{figure}
    \centering
    \includegraphics[width=0.7\linewidth]{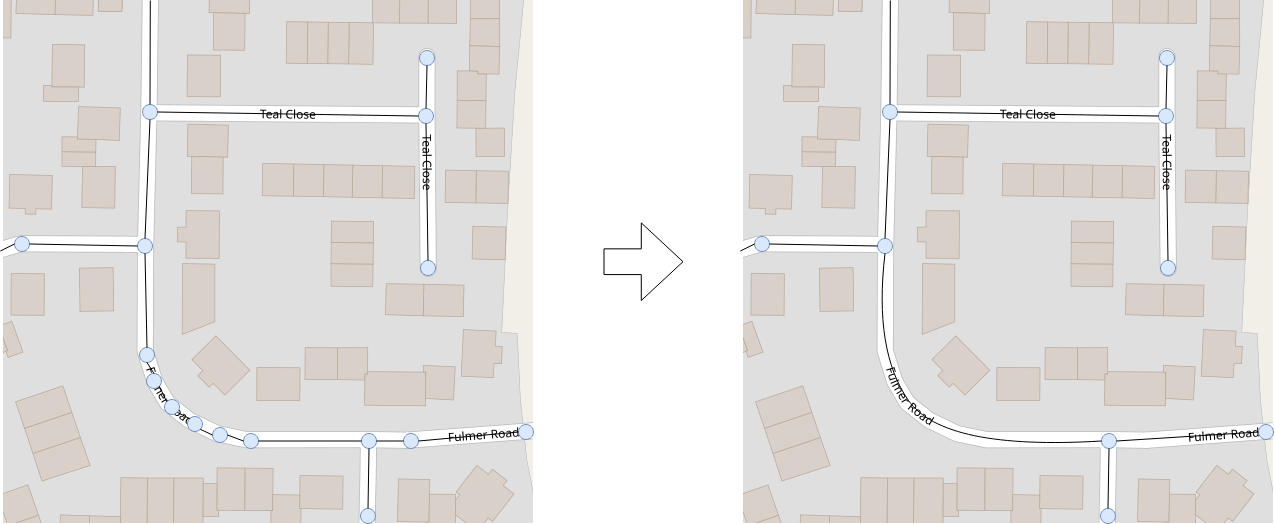}
    \caption{The left side shows a road graph that could be in the OpenStreetMap dataset. The curved street is represented by many additional nodes to get a visually accurate representation. As we are more interested in a logical graph representation, we can remove such degree two nodes to derive a compacted graph which is illustrated on the right.}
    \label{fig:compaction}
\end{figure}
The graph structure for the competition is based on OpenStreetMap data. Primarily used for visualizing the graph, it is built so that every edge can be drawn with a single straight line segment. 
Curved roads have to be represented with additional nodes to display the turn as depicted in Figure~\ref{fig:compaction}.
As we use message passing GNNs that propagate information one hop per layer, bends in roads could lead to bottlenecks that unnecessarily hinder the exchange of information.
Therefore, we remove as many of these nodes as possible.
We do this by (1) identifying nodes in the directed graph with precisely two neighbors with directed edges from and to both. Additionally, we check whether the two neighbors are connected to prevent introducing multi-edges into the graph.
In step (2), we replace the node with a single edge. We recompute updated features for the new edge based on the features of the two replaced edges: \texttt{parsed\_maxspeed} and \texttt{importance} are averaged while \texttt{length\_meters} is summed up.
To preserve crucial traffic counter information, we copy the values of available traffic counts to the neighbors, given that they already do not have traffic counts for the time interval.

We further compute new ground truth labels by combining the labels of the two replaced edges. During training, we only worked on the compacted graph. For inference, we keep a mapping from edges in the compacted graph to edges in the original graph. The model then predicts congestion classes on the compacted graph and uses the prediction of an edge for all edges that were replaced by it. 
Lastly, we also remove isolated nodes. Figure~\ref{fig:two_tables} shows the sizes for the graphs before and after compaction.

\section{Supersegment Graph Extensions}
\label{sec:supersegment}
\begin{figure}
    \centering
    \includegraphics[width=\linewidth]{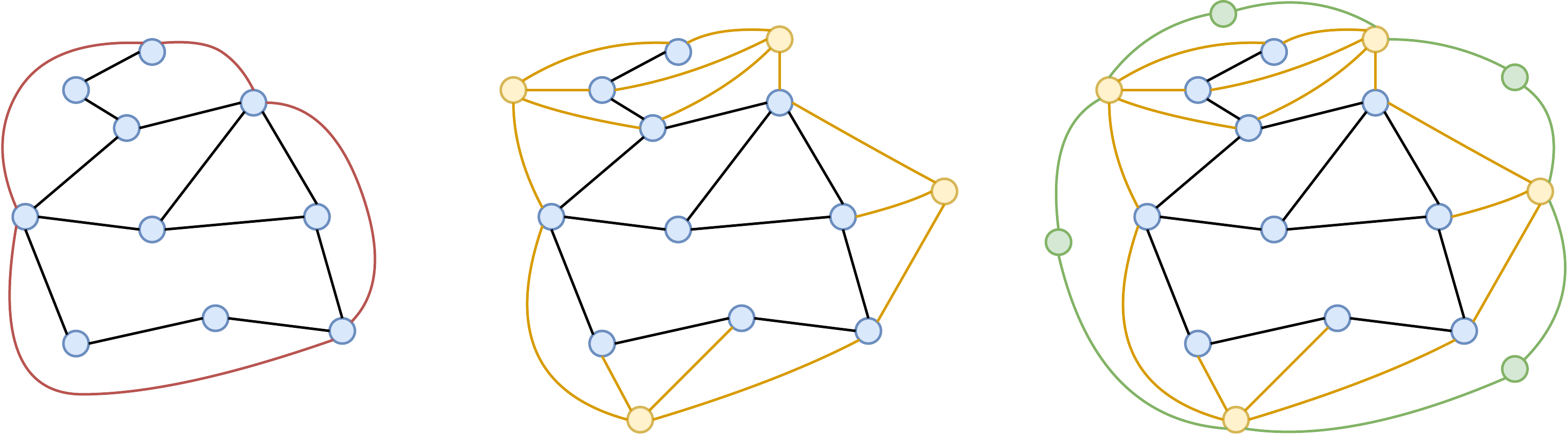}
    \caption{The three hierarchical graph representations to model the supersegments. From left to right: (1) Adding red edges between supernodes, (2) Adding additional yellow nodes for supersegments and connecting them to the contained nodes and (3) also add green nodes for edges in the supersegment graph. }
    \label{fig:supergraphs}
\end{figure}
For the extended competition, a set of supersegments was provided for every city graph. To construct the supersegment graph, important nodes were identified in the original graphs (we will call them ``supernodes'' from here on) and connected to their k nearest neighbors based on the shortest path distances as described in the official documentation of the extended challenge.  
One supersegment then corresponds to an edge between two supernodes and contains all nodes that lie on the shortest path between the two supernodes in the original graph.
We used the supersegments provided and added additional edges or nodes to the original graph to simplify the message exchange between important nodes.
We tested three approaches depicted in Figure~\ref{fig:supergraphs}. 

In version (1), we add edges between supernodes that are connected to each other in the supersegment graph. 
Between message passing steps on the original graph (using GNNConv), we interleave layers every two steps that only exchange messages along these additional edges (using GNNPoolConv).
This reduces the graph's diameter and supports the information exchange between ``important'' intersections.

In version (2), we first pool information from all nodes lying on the same supersegment and then propagate it back to them.
We add one node per supersegment to the original graph (called ``supersegment nodes'' from here on) and connect it to all nodes contained in that supersegment (using directed edges).
First, we execute four GNNConv convolutions on the original graph, followed by a pooling step using the supersegment nodes. Then, we execute an additional two GNNConv convolutions on the original graph. 
To pool the embeddings, we apply two GNN layers (GNNPoolConv). The first layer propagates messages on the directed edges to the supersegment nodes. The other layer propagates messages the other way (from supersegment nodes to the nodes in the original graph).
This enables nodes to share contextual information within the same supersegment.
The supersegment nodes can also be used to make predictions for the extended competition.
The message passing to the additional nodes is done for every 3 iterations of normal message passing.

As an extension of the second approach, in (3), we add nodes for every node in the supersegment graph and connect them to the supersegment nodes.
The model then propagates messages to the supersegment nodes as before. Before propagating the messages back to the original graph nodes, two layers of message passing are executed on the edges between the supersegment nodes and the additional nodes.
Compared to the second approach, this enables information to be exchanged not just on the same supersegment but also between supersegments that overlap.

Figure ~\ref{fig:two_tables} shows the sizes of the resulting graphs built with the three different approaches.

\section{Node and Edge Features}
Besides processing the graph structure, we select features for the nodes and the edges and compute global graph statistics, which we add to every node. We use the same features for both the core and extended part of the competition.
Every node has 14 features. The first two consist of the x and y positions of the node in the graph given by the GPS coordinates. Both are normalized to be within the interval $[0,1]$. The following four features are the normalized loop counts for the past four 15 minutes intervals at the specific node. Furthermore, we compute two simple global statistics over all loop counts in the graph. So for each of the 4 loop counts, we compute the mean and standard deviation as global statistics and add them as node features.
For each edge, we compute four features. The first three consist of normalized \texttt{parsed\_maxspeed}, \texttt{importance}, \texttt{length} as specified by the OpenStreetMap data. Furthermore, we introduce a fourth feature defined as \texttt{length} divided by \texttt{parsed\_maxspeed}. The features that were not available, i.e., NaN values for the counts, were ignored to compute the global statistics and were replaced with a default value of -1. We do not explicitly address the sparsity of the loop counter in the data preprocessing phase. We instead rely on the graph convolutions to exchange the sparse data sensibly among neighbors. However, note that every node (even if it has no counts) has access to the global statistics features. 
The node and edge features are then encoded with a VertexEncoder and EdgeEncoder as illustrated in Figure \ref{fig:model Description}.

\begin{figure}%
    \centering
    \subfloat[\centering The graph compaction can reduce the graph size up to thirty percent by removing degree two nodes. ]{{
  \begin{tabular}{llll}
    \toprule
    \cmidrule(r){1-2}
    Graph           & City          & Nodes         & Edges \\
    \midrule
                    & London        &  59110    & 132414\\
    Original Graph  & Madrid        &  63397    & 121902\\
                    & Melbourne     &  49510    & 94871\\
    \midrule
                     & London        &  39762    & 93718\\
    Compact graph    & Madrid        &  60366    & 115840\\
                     & Melbourne     &  39035    & 73921\\   
    \bottomrule
    \vspace{0.95cm}
  \end{tabular}
    }}%
    \subfloat[\centering Three different approaches to model the supersegments. The table denotes the number of additional nodes and edges added to the original graph.]{{
    \begin{tabular}{llll}
    \toprule
    \cmidrule(r){1-2}
    Graph           & City          & Nodes         & Edges \\
    \midrule
                    & London        &  -    & 8024\\
    Approach (1)   & Madrid        &  -    & 7938\\
                    & Melbourne     &  -    & 6492\\
    \midrule 
                    & London        &  4012    & 118999\\
    Approach (2)   & Madrid        &  3969    & 102456\\
                    & Melbourne     &  3246    & 82620\\   
    \midrule
                    & London        &  4512    & 135047\\
    Approach (3)      & Madrid        &  4431    & 118332\\
                    & Melbourne     &  3692    & 95604\\
    \bottomrule 
  \end{tabular}
    }}%
    \caption{Comparison of different graph representations of the street networks for the specific cities. The left table illustrates the size difference between the original OpenStreetMap graph and the compacted version. The right table summarizes the sizes of the different approaches to model the supersegments.}%
    \label{fig:two_tables}%
\end{figure}


\section{Training and Model Selection}

\begin{figure}[h]
    \centering
    \includegraphics[width=1.0\linewidth]{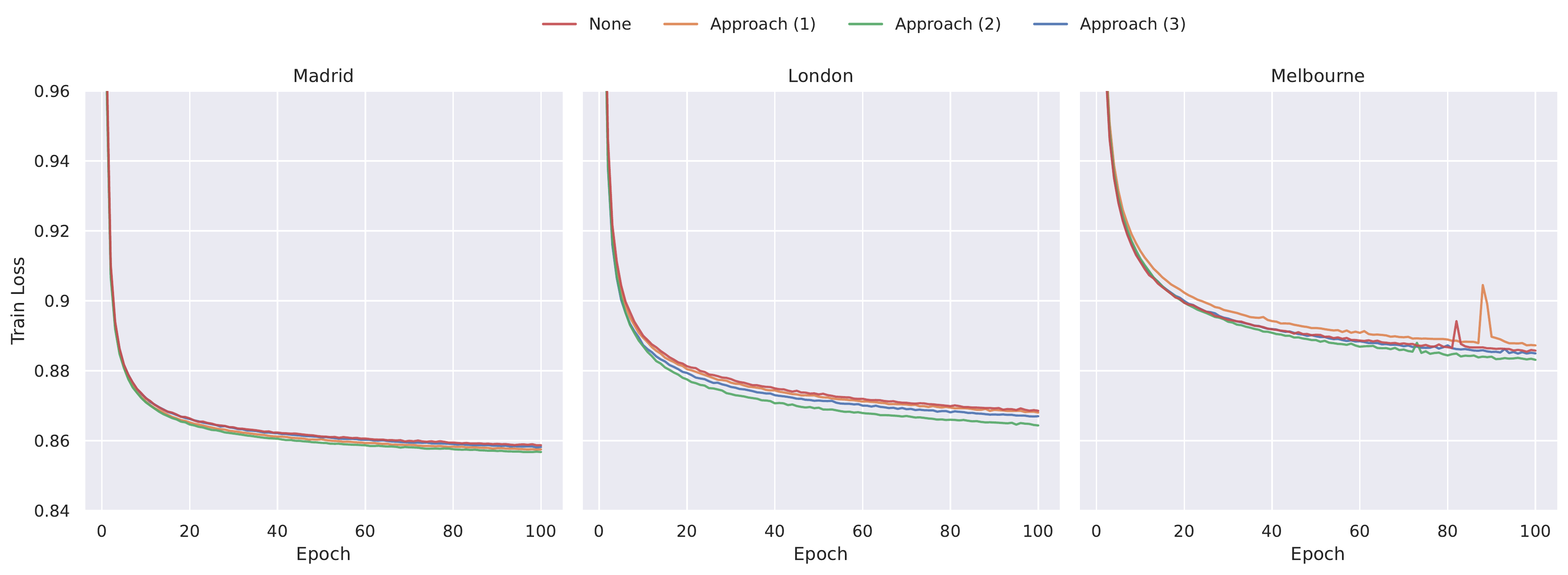}
    \caption{Comparison of the different supergraph approaches. Approach (2) consistently yields better results across all cities compared to the other supergraphs or only using the original graph topology. The runs were done on uncompacted graphs.}    \label{fig:supergraph_losses}
\end{figure}

We trained separate models for each city and task (core and extended). 
Due to a shortage of training time for the final runs, we could not extensively compare all of our variants for the models we submitted.
Furthermore, the final models were trained without graph compaction.
We started training runs with a train/validation split of 80\%/20\% to observe our models' quality. 
Before the final submission deadline, we decided to continue some trainings on the full dataset by loading checkpoints from old runs using the split and continuing the training on the full dataset.
These runs ended up being the best for the core competition. The exact epochs trained on the different splits are contained in Table~\ref{tab:models}.
For the extended competition, we did not use the runs on the full dataset for our submission.
We got the best test score by taking the according models (for each city) that achieved the best validation score when trained on the 80\%/20\% split.
Training data was always shuffled, and due to the big memory footprint during training, we were limited to a batch size of 1.
Our GNN architectures were implemented by using the PyG framework \citep{pyg}.
The AdamW optimizer was used for training with a learning rate of 5e-4 and weight decay set to 1e-3. 
The models for the core competition are trained on the weighted cross-entropy loss, whereas the models for the extended competition are both trained on a weighted combination of the L1 loss of the eta and the cross-entropy loss for the congestion classes simultaneously. 
All training runs were performed on a RTX 3090 with 24 GB of VRAM. 

During development, we experimented with the compacted graph representation, which seemed to work well while using less resources. However, for our main submissions, we focused on adding a hierarchical graph representation. We used the original graph in combination with Approach (2) outlined in Section \ref{sec:supersegment} for the core task and Approach (1) and (2) (added edges between supernodes and pooling on the supersegments with separate message passing rounds) for the extended competition. 

Furthermore, we compare the different approaches in Figure \ref{fig:supergraph_losses}. While all supergraph segments improve performance in comparison to using the original graph topology only, Approach (2) seems to perform best.

\section{Discussion}

We used a Graph Neural Networks centered approach to participate in the Traffic4cast 2022 edition. The primary motivation is to easily incorporate both the given input data and predictions, which were specified on the road network topology of the original OpenStreetMap graph and further process the original graph to enable better information flow.  
We observe that the OpenStreetMap graph can be compacted as it contains additional artifacts to visualize the graphs well. This could be an interesting future direction to combine with the supersegment representation. Furthermore, we outline three possible approaches to model the combined graph, including the supersegments. Our final approach focuses on a combined hierarchical representation of supersegments to strengthen the information flow between key intersections of the original road topology. 
Additionally, we use a multi-task setting for the extended competition where we train on predicting the ETAs and congestion classes simultaneously.
Lastly, we see much potential in exploring different ways of constructing and defining supersegments, which could allow for multiple levels of abstraction. 

\begin{landscape}
\begin{table}[p]
  \centering
  \begin{tabular}{llllllllll}
    \toprule
    \cmidrule(r){1-2}
    Model (checkpoint)           & Architecture          & \makecell{Fine-tuning\\Freezing}         & Sampling & \makecell{Size \\training set} & Training data & \makecell{Number of\\ iterations (epochs)} & Batch size & \makecell{Trainable\\ parameters} & Optimizer\\
    \midrule
    \makecell{Supersegment \\ Approach (2)}   & \makecell{Traffic \\ Core CC}        &  -    & \makecell{80\% (then 99\%)\\ per city, \\6:00 - 22:00} & First 99\% & \makecell{(14,) per node,\\ (4,) per edge} & \makecell{54+19 (London),\\ 76+30 (Melbourne),\\ 55+22 (Madrid)} & 1 & 4,936,964 & AdamW\\
    \midrule    \makecell{Supersegment \\ Approach (1) + (2)}   & \makecell{Traffic \\ Extended ETA}       &  -    & \makecell{80\%\\ per city, \\6:00 - 22:00} & First 80\% & \makecell{(14,) per node,\\ (4,) per edge} & \makecell{33 (London),\\ 41 (Melbourne),\\ 24 (Madrid)} & 1 & 5,134,340 & AdamW\\

    \bottomrule
  \end{tabular}
  \caption{Configuration of the training setup for our chosen architectures. The number of epochs X+Y is to indicate that first, the model is trained X epochs on 80\% of the data and then trained for additional Y epochs on 99\% of the dataset.}
    \label{tab:models}
\end{table}
\begin{figure}[p]
    \centering
    \includegraphics[width=0.7\linewidth]{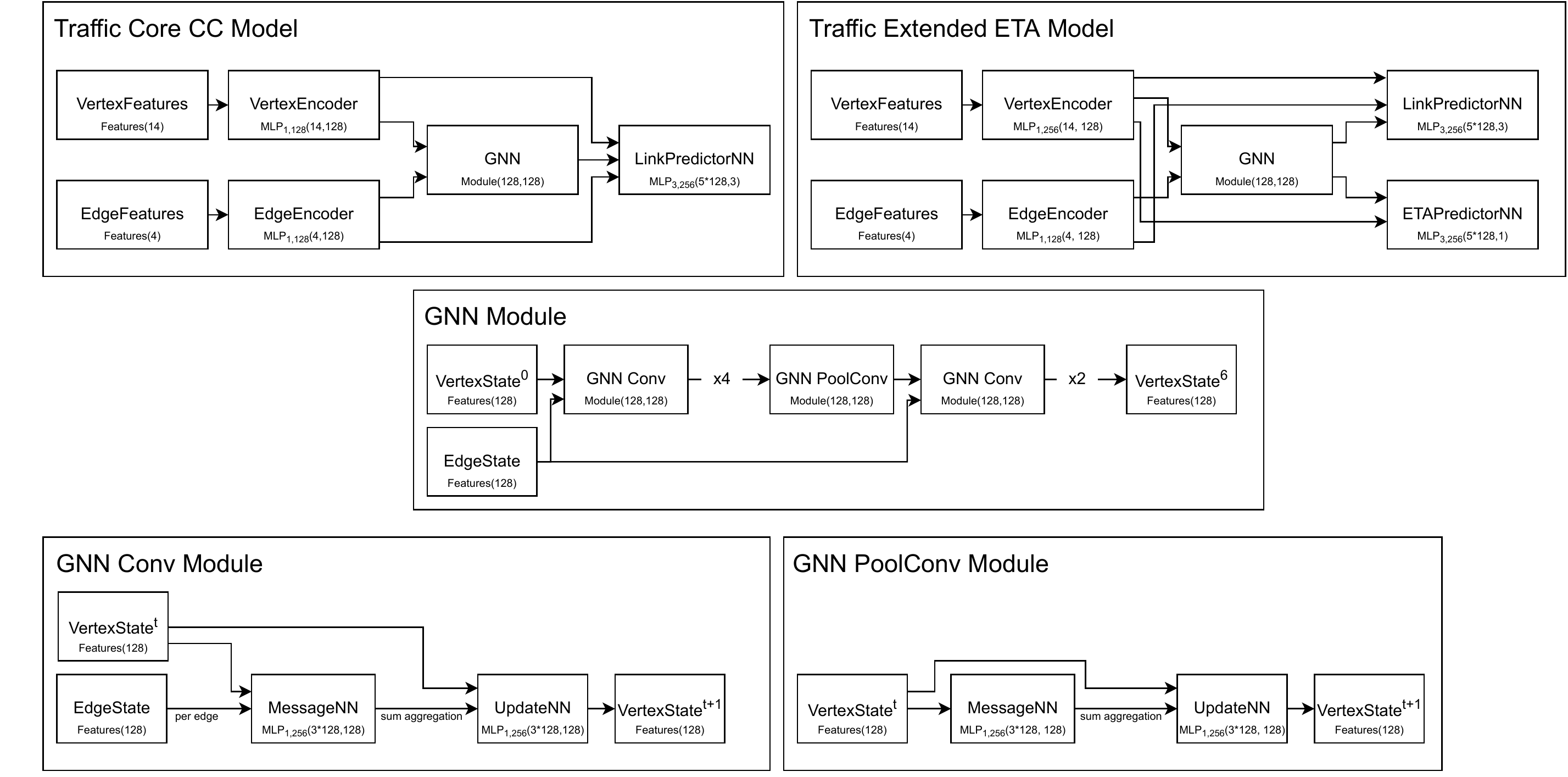}
    \caption{Overview of our model architecture. The nodes and edges are encoded separately into an embedding space. The GNN module consists of multiple rounds of graph convolutions. Each graph convolution executes a single step of message passing. The final predictors take the node and edge embeddings as well as the embeddings after the GNN to predict the congestion classes or ETAs. MLP$_{d,s}(in,out)$ denotes an MLP of input dimension $in$, depth $d$ with hidden dimension $s$ and output dimension $out$. }
    \label{fig:model Description}
\end{figure}
\end{landscape}






\bibliographystyle{plainnat}
\bibliography{refs.bib}

\end{document}